\documentclass[sigconf,nonacm]{acmart}

\usepackage{amsmath}
\usepackage[ruled,vlined,noend]{algorithm2e}

\usepackage{multirow}
\usepackage[switch]{lineno} 
\usepackage{float}

\usepackage{colortbl}
\newcommand{\todo}[1]{\textcolor{red}{#1}}
\newcommand{\OC}[1]{\textcolor{black}{#1}}
\newcommand{\FT}[1]{\textcolor{black}{#1}}
\newcommand{\FTT}[1]{\textcolor{black}{{#1}}}
\newcommand{\JO}[1]{\textcolor{black}{#1}}

\newcommand{\JNEW}[1]{\textcolor{black}{#1}}

\newcommand{\FTd}[1]{\textcolor{black}{#1}}
\newcommand{\deld}[1]{
}
\newcommand{\nosp}[1]{
}
\begin{document}

\title{Automatic Product Ontology Extraction from Textual Reviews}

\author{Joel Oksanen}
\affiliation{%
  \institution{Imperial College London}
  \city{London}
  \country{UK}
}

\author{Oana Cocarascu}
\affiliation{%
  \institution{King's College London}
  \city{London}
  \country{UK}
}

\author{Francesca Toni}
\affiliation{%
  \institution{Imperial College London}
  \city{London}
  \country{UK}
}

\begin{abstract} 
Ontologies 
have proven beneficial in different settings that make use of textual reviews\nosp{, ranging from sentiment analysis to e-commerce and recommender systems}. 
However,
manually constructing ontologies is a laborious and time-consuming process in need of  automation. 
%
We propose a novel methodology for automatically extracting ontologies\FTd{, in the form of meronomies,} from product reviews\FTd{,} \JNEW{using \FTd{a very limited amount of} hand-annotated training data}. \nosp{Our method uses masked BERT within a distantly supervised learning setting
to identify aspects and 
\FTd{{\em part of}} relations between them.} We show that the ontologies generated by our method outperform hand-crafted ontologies (WordNet) and  ontologies extracted by  existing  methods (Text2Onto \FTT{and COMET}) in \FTd{several, diverse} settings. Specifically, 
\FTd{our generated} ontologies \FTd{outperform the others when evaluated by} 
human annotators as well as on an existing Q\&A dataset from Amazon\FTd{. Moreover, our method is  better able to generalise, in  capturing 
knowledge about unseen products. Finally, we consider } 
a real-world setting, \FTd{ showing that our method is better able} to determine recommended products based on their reviews, 
in alternative to using Amazon's standard score aggregations. 
\deld{Finally, we evaluate the ability of our method to generalise in order to capture 
knowledge about unseen products.}
\end{abstract}

\maketitle

\section{Introduction}

Ontologies have proven beneficial in a variety of settings that operate with textual reviews, such as sentiment analysis \FT{(e.g., as in \cite{HemmatianS19,ZhouHCHWH19})}, e-commerce \FT{(e.g., as in \cite{KaramanolakisMD20})}, and recommender systems \FT{(e.g., as in \cite{DauSRO20})}. \FT{In particular, p}roduct ontologies can help users 
quickly obtain information about 
products and thus aid with 
purchase decision making. 
\FT{However,} constructing ontologies manually is a 
\FT{laborious and time-consuming} process 
\FT{in need of} automation. In recent years, several approaches have been proposed to automatically construct ontologies (for a  \FT{recent} survey 
see \cite{RefWorks:doc:5f380d63e4b088f8a2f1f066}).

Different types of relations can be used \nosp{to characterise} \FTd{in} ontologies. \nosp{In this paper, we} \FTd{We} focus on extracting \textit{meronomies} that represent \textit{part of} relations where a node's children are constituent parts of that node (e.g., \FT{the} parent \FT{may be} \textit{camera} and \FT{one of its} child\FT{ren} \FT{may be} \textit{lens}).
%
\FT{Alternatively, relations may amount to} \emph{taxonomies} (e.g., \FT{as in}  Graph2Taxo
\cite{RefWorks:doc:5f380b18e4b0f05471536f61}) 
\FT{amounting to} \textit{is a} relations where a node's children are more specialised forms of that node (e.g., the parent may be \textit{camera} and one of its children may be \textit{digital camera}). 
In this paper, the taxonomy of a node is included within that node's \textit{synset} \FT{(sets of synonyms)}, i.e.\FTd{,} we do not distinguish between the different forms \FT{aspects (i.e., \emph{products}, their \emph{features}, \emph{sub-features} etc.) may take} (e.g., \textit{film camera} and \textit{digital camera} \FT{are all within the synset for \emph{camera}}).

\FT{We} propose a \FTd{novel} method for extracting \FT{product} ontologies \FT{(of the form just described)} from  textual product reviews, using \emph{masked BERT}  \cite{RefWorks:doc:5e8dcdc3e4b0dba02bfdfa80} \FT{within a \emph{distantly supervised} setting}.
%
Specifically, \FTd{our method incorporates} a BERT model for identifying the various \emph{aspects} of 
\FT{products} and a\FT{nother} BERT model to determine \emph{sub-feature relations} between aspects. In addition, we train word2vec \cite{RefWorks:doc:5edbafdde4b0482c79eb8d95} to extract word vectors which are used to group the aspects into 
\emph{synsets}
using the method proposed in \cite{RefWorks:doc:5eaebe76e4b098fe9e0217c2}. 
%
%
Our method \FTd{requires a very limited amount of hand-annotation, relying}  on a 
\FTd{very} small, manually created ontology and \FTd{using} 
distantly supervised learning where 
a large amount of text 
\FT{is \emph{automatically} annotated}
.

We evaluate our method on product reviews from Amazon\footnote{\url{https://nijianmo.github.io/amazon/index.html}} with several, diverse 
experiments showing 
that \FTT{an ontology} 
 extracted by our method outperforms 
a lexical database of semantic relations between words, WordNet \cite{RefWorks:doc:5f46c7b9e4b0bb83a698cd37}, and 
\FTT{two} ontology learning framework\FTT{s}, Text2Onto \cite{RefWorks:doc:5f46c8aae4b08553d3e6b982} 
\FTT{and COMET} \JO{\cite{Bosselut:19}}. \FT{ The experiments amount to\nosp{ the following}:} \textit{1)}  ontology evaluation with human annotators, \textit{2)} ontology evaluation on a Q\&A dataset on Amazon products \cite{McAuleyY16,WanM16}, \textit{3)} 
ability 
to generalise on unseen products.
We also evaluate the usefulness of  our method  \FTd{in a downstream task,} to produce ontologies amenable to
determine recommended products based on their reviews as an alternative to the aggregated scores shown on \nosp{the} Amazon\nosp{ website}\footnote{\url{https://www.amazon.com/}}.

%

\section{Related work} \label{sec:related}


Several works have proposed methods for ontology construction \nosp{in the} \FTd{for} e-commerce\nosp{ domain}\FT{, e.g., } \cite{RefWorks:doc:5f38122be4b0a8a5dc742bcd,Konjengbam:18,KaramanolakisMD20}\JNEW{, the vast majority \nosp{of which focus} \FTd{focusing} on taxonomies.} 
\cite{KaramanolakisMD20} propose a deep neural network that leverages the 
taxonomy of product categories and extracts attributes conditional to its category. The method is evaluated on a taxonomy of 4000 product categories.
Other works 
\FT{also focus on taxonomies} \cite{LuoLYBCWLYZ20, Shang:20, Shen:20, Kozareva:10}.
\JNEW{Product ontology extraction in the form of meronomies is \deld{more overlooked} \FTd{much less explored}\nosp{  in the literature}.}
\FTd{Amongst these,} \cite{Konjengbam:18} use Latent Semantic Analysis to represent words as vectors in high-dimensional space and cosine similarity to construct \deld{the aspect ontology} \FTd{aspect ontologies, testing their method} 
on four tech-related product types: digital camera, mobile phone, DVD player, and mp3 player.
%
%
\FTd{Further,} \cite{RefWorks:doc:5f38122be4b0a8a5dc742bcd} extract ontologies from product reviews \FTd{using structured input metadata and predefined relations between aspects\nosp{ for each product},} and test their method on two product categories only: smartphones and digital cameras.
\deld{Their method relies on structured input metadata and predefined relations between aspects for each product.}

\JNEW{Huge semantic knowledge bases, such as YAGO \cite{Tanon:20}, DBpedia \cite{Auer:07}, and ConceptNet \cite{Speer:17}, are \FTd{a} popular source for structured common knowledge. However, their knowledge is often limited when it comes to niche categories such as specific products: for example, the \textit{HasA} relation for \textit{kettle}\footnote{\url{https://conceptnet.io/c/en/kettle}} on English ConceptNet gives only one result of \textit{handle}, missing important features such as \textit{base}, \textit{switch}, and \textit{material}. 
Therefore, such knowledge bases are unsuitable for constructing a useful and comprehensive product ontology.}

\FT{Masking  has built-in support 
in BERT \FTT{\cite{RefWorks:doc:5e8dcdc3e4b0dba02bfdfa80}}, where it
plays a central role in 
the pre-training phase. Indeed, one of the tasks BERT is trained on is \textit{masked language modelling} where random words are replaced with a [MASK] token and the model needs to predict these words based on the context provided by the non-masked words in the sequence.}
\cite{MalaviyaBBC20} showed that BERT performs well at capturing
\textit{made of} and \textit{part of} relations,
due to masking tokens during pre-training. 
\cite{RefWorks:doc:5edc9ecbe4b03b813c4d4381} explored the use of masked language models by masking fine-grained aspects during training on review data from Yelp and Amazon clothing.
There are other works that use deep learning techniques to extract ontologies \cite{Navarro-AlmanzaJLC20} and knowledge graphs \cite{LiuZ0WJD020} but on different types of data and tasks than the one addressed in this paper.
\FTT{In this paper we will evaluate our method against two existing methods which can be used to \JNEW{automatically} extract ontologies in the form of meronomies \JNEW{from text}: Text2Onto \cite{RefWorks:doc:5f46c8aae4b08553d3e6b982}  and COMET \JO{\cite{Bosselut:19}}} \FTT{used with ConceptNet}.

There are also works in the area of aspect-based sentiment analysis that involve the identification of aspects in reviews (for an overview of methods used in this area see surveys \cite{HemmatianS19,ZhouHCHWH19}). Whilst our aim is not on performing sentiment analysis, the aspects extracted for this task can be used to create an ontology.


\section{Methodology} \label{sec:method}

\begin{figure*}
	\centering
	\includegraphics[width=0.9\textwidth]{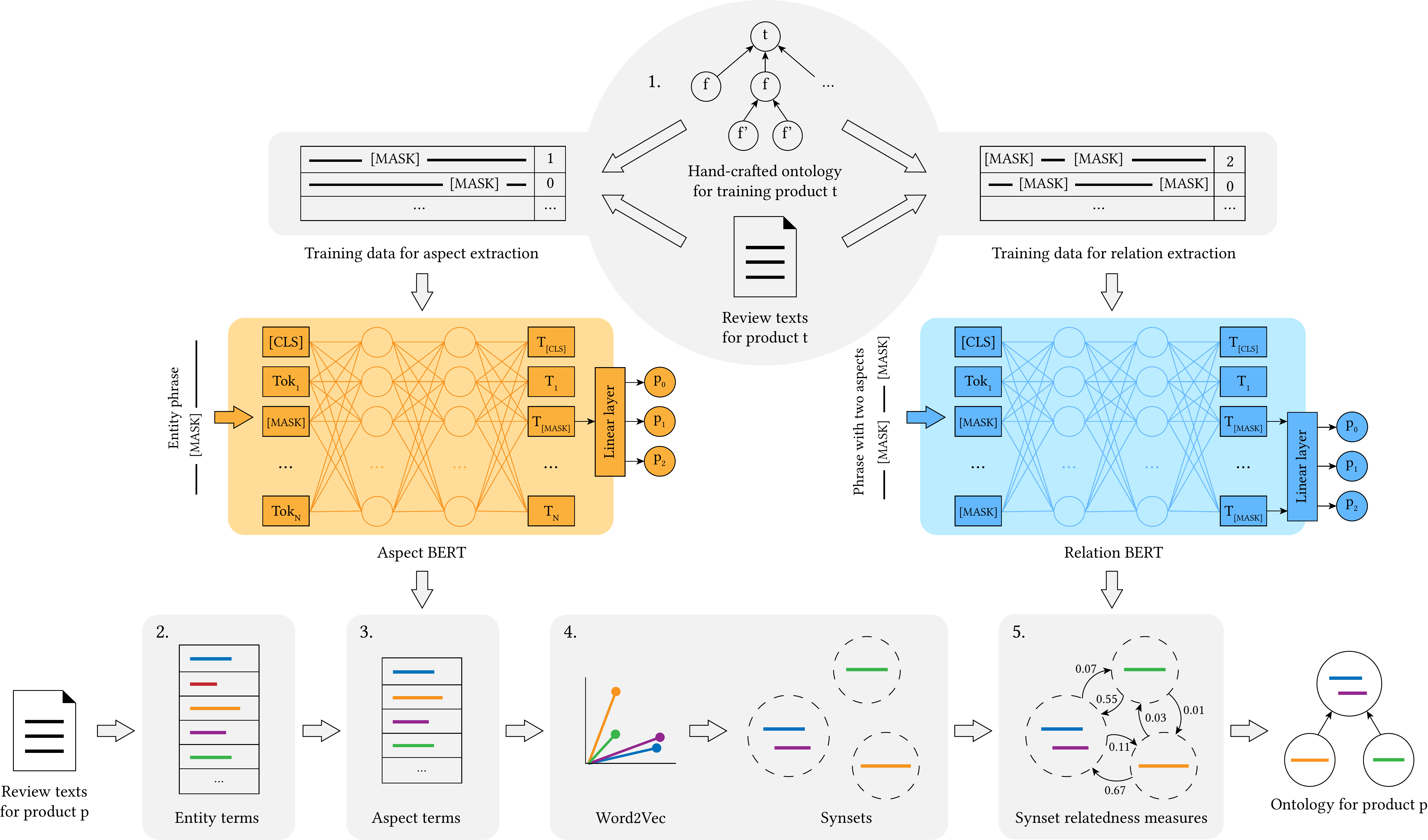}
	\caption{An overview of our methodology. The labels 1 to 5 refer to the corresponding subsections: \ref{sec:annotation}) annotating data, \ref{sec:noun_extraction}) 
	entity extraction, \ref{sec:feature_extraction}
	) aspect extraction, \ref{sec:synonym}) synonym extraction, and \ref{sec:ontology_extraction}) ontology extraction. Here, in the Aspect BERT on the left, p$_0$, p$_1$, p$_2$ stand, respectively, for the probability of  the entity phrase in input being a non-aspect, a feature aspect, or a product aspect. In the Relation BERT on the right, p$_0$, p$_1$, p$_2$ stand, respectively, for the probability that the two input aspects in the input phrase are not related,  the second aspect is a feature of the first, or  the first aspect is a feature of the second.}
	\label{fig:pipeline}
\end{figure*}

We \FT{first} present \FT{(Section~\ref{sec:annotation})} the annotation method \FT{we use} 
to obtain \FT{automatically, from a small hand-crafted ontology,} the training data for our two BERT models,  and \FT{then} describe 
\FT{our} 
four-stage 
pipeline for extracting ontologies from text:
entity extraction \FT{(Section~\ref{sec:noun_extraction})}, aspect extraction \FT{(Section~\ref{sec:feature_extraction})}, synonym extraction \FT{(Section~\ref{sec:synonym})}, and ontology extraction \FT{(Section~\ref{sec:ontology_extraction})}, \FT{where aspects amount to products or features, and
ontology extraction includes a component for extracting relations between aspects.} Figure~ \ref{fig:pipeline} shows an overview of the pipeline 
underpinning our  methodology.

\subsection{Annotating Training Data for Masked BERT}
\label{sec:annotation}

Annotating 
\FT{reviews with ontological information} that would be representative of 
\FT{an entire set of products (such as available on Amazon)} would be nearly impossible due to the large number of different product categories. However, \FT{we can benefit from the observation that,} in review texts, certain grammatical constructs are similar regardless of the product. 
Consider the following two sentences, where \textit{lens} is a feature of \textit{camera} and \textit{material} is a feature of \textit{sweater}: \begin{center}
    \textit{I love the \textbf{lens} on this \textbf{camera}}

    \textit{I love the \textbf{material} of this \textbf{sweater}}
\end{center}
Masking the entities 
\FT{in bold} in the sentences, we obtain:
\begin{center}
\textit{I love the \textbf{e1} on this \textbf{e2}} \quad and \quad \textit{I love the \textbf{e1} of this \textbf{e2}}
\end{center}
The two sentences are nearly identical. 
\FT{Indeed, w}
hile the entities in these review texts are domain-specific, the context is  domain-independent.
\FT{Then, if we know that \textit{lens} is a feature of product \textit{camera}, with masking}, we can train a classifier to recognise that \FT{\textit{material} is a feature of product \textit{sweater}.}
\FT{This} example 
 is, 
\FT{clearly},
an idealised scenario\FT{, but similar considerations can be made more generally. For example,}  consider the following two sentences:
\begin{center}
    \textit{The \textbf{camera housing} is made of shiny black \textbf{plastic} but it feels nicely weighted 
    } 

    \textit{It's a lovely warm \textbf{jumper} that has a nice \textbf{feel} to it}
\end{center}
Masking the entities 
\FT{in bold}, we obtain:
\begin{center}
    \textit{The \textbf{e1} is made of shiny black \textbf{e2} but it feels nicely weighted
    }

    \textit{It's a lovely warm \textbf{e1} that has a nice \textbf{e2} to it}
\end{center}
\FT{These} 
still contain some rather domain-specific terms such as \textit{shiny}, \textit{weighted}, and \textit{warm}
\FT{, but they} 
are 
\FT{nonetheless} common to wider categories of products, such as \textit{electronics} or \textit{clothing}. 
Thus, we 
hypothesise that a 
varied set of \FT{annotated} reviews  for  \FT{few} selected products should 
\FT{suffice} to train domain-independent classifiers \FT{for ontology extraction}.
\FT{However, m}
anually annotating reviews for 
\FT{even a handful of} products would still require 
\FT{looking at} several thousands of sentences in order to obtain a sufficiently large dataset \FT{for training}.  
   \FT{Our method uses} \textit{distantly supervised learning} \FT{with automatically annotated reviews from}
%
\FT{a manually created, small}
ontology for a few (randomly) chosen categories of products\footnote{\FT{In the experiments, we use an ontology with just 5 products,  maximum number of (sub-)features per product 54 and maximum depth (features, sub-features, etc.) 5, which took less than
2 hours to create.}}. \FT{An} automated process  search\FT{es} 
review texts for terms that appear in 
\FT{this} ontology, and \FT{masks them while also labelling sentences in the review texts, as indicated below, to obtain training data for the two tasks of aspect and relation extraction.}
%
Note that 
without masking 
the classifier\FT{s} would simply learn to recognise the terms in the \FT{manually created} ontology
\FT{; instead,} 
with masking, 
they are
forced to rely solely on 
\FT{the terms' context,} which is more varied.

To obtain training data for the aspect extraction task, 
\FT{our method automatically selects} review sentences containing exactly one commonly appearing entity in the reviews and \FT{
labels these sentences with: }  0 
\FT{if the entity is} not an aspect, 1 
\FT{if the entity is} a feature aspect, or 2 
\FT{if the entity is} a product aspect, as 
\FT{illustrated} in Table \ref{tab:fe_training_data}. Selecting only a subset of the sentences 
\FT{allows} us to reduce 
the aspect recognition task (see Section~\ref{sec:feature_extraction})
to a classification task instead of a sequence-labelling problem. \FT{At the same time,} 
even relatively small product categories have large amounts of review data available
\FT{, so 
pruning the number of 
sentences 
does not jeopardise our ability to obtain a sufficient 
amount of data.}

\begin{table}
	\centering
	\caption{Training data examples for aspect extraction with entities highlighted in bold.}
 	\begin{tabular}{l c}
 	\multicolumn{1}{c}{Text} & Label \\
 	\hline
 	``My \textbf{daughter} loves it!" & 0 \\
 	``The \textbf{material} is super soft." & 1 \\ 
 	``I love this \textbf{sweater}." & 2 \\ 
	\end{tabular}
	\label{tab:fe_training_data}
\end{table}

\begin{table}
	\centering
	\caption{Training data examples for relation extraction with entities highlighted in bold.}
	\begin{tabular}{l c}
 	\multicolumn{1}{c}{Text} & Label \\
 	\hline
 	``I like the \textbf{design} and the \textbf{material}." & 0 \\ 
 	``The \textbf{sweater}'s \textbf{fabric} is so soft." & 1 \\
 	``The \textbf{colour} of the \textbf{sweater} is beautiful." & 2 \\
	\end{tabular}
	\label{tab:re_training_data}
\end{table}

\FT{To obtain data for the relation extraction task,} 
\FT{our method automatically selects} review sentences containing exactly two aspects from the \FT{manually} annotated ontology and labels 
\FT{each such} sentence as follows: 1 if the second aspect is a feature of the first aspect, 2 if the first aspect is a feature of the second aspect, and 0 otherwise, as 
\FT{illustrated} in Table \ref{tab:re_training_data}. \FT{Note that a}n aspect $a_1$ is considered a feature of aspect $a_2$ if and only if $a_1$ is a descendant of $a_2$ in the annotated ontology tree. For example, \textit{fabric} is considered a feature of both \textit{sweater} and \textit{material}\FT{, where in the ontology tree  \textit{sweater} is the parent of \textit{material} which in turn is the parent of \textit{fabric}}. 
\FT{The training data for the models we  experiment with in Section~\ref{sec:evaluation} are drawn (as presented above) from a manually created input ontology obtained as follows. }
We select 5 random product categories from
\cite{RefWorks:doc:5edc9ecbe4b03b813c4d4381}: \textit{digital cameras}, \textit{backpacks}, \textit{laptops}, \textit{acoustic guitars}, and \textit{cardigans}. For each of these categories, \FT{
we identify  the 200 most commonly appearing nouns in the review texts, and manually create the (input) ontology around them, given that} 
we observed that most of the relevant aspects of the products 
\FT{are} included within this \FT{(200 nouns)} range. 
Table \ref{tab:training_data} shows 
statistics \FT{for the two datasets (of automatically labelled reviews, by distant supervision from this 5-products input ontology)} obtained after undersampling the over-represented classes 
to 
\FT{guarantee that the datasets are} balanced
.

\begin{table}
	\centering
	\caption{Number of training instances for the two classifiers.}
 	\begin{tabular}{l c c}
 	Classifier & Per product & Total \\
 	\hline
 	Aspect extraction & 42,313 & 211,565 \\ 
 	Relation extraction & 21,131 & 105,655 \\ 
	\end{tabular}
	\label{tab:training_data}
\end{table}

\subsection{Entity Extraction}
\label{sec:noun_extraction}

The first stage of our ontology extraction method is to extract the most commonly appearing entities in the review texts. We split the review texts into sentences, treating each sentence as an individual unit of information independent from other sentences in the same review text. We then tokenise the sentences and use an out-of-the-box implementation of \cite{RefWorks:doc:5edca760e4b0ef3565a5f38d} to join common co-occurrences of tokens into bigrams and trigrams. This step is crucial to detect multi-word entities such as \textit{operating system}, which is an important feature of \textit{computer}. We then use the part-of-speech tagger from NLTK\footnote{\url{https://www.nltk.org/}} to select the noun entities within the tokens and count the number of occurrences for each of the entities across all of the review texts. Finally, 
the 200 most commonly appearing entities 
\JNEW{are passed} onto the aspect extraction 
\FT{stage}.

\subsection{Aspect Extraction} \label{sec:feature_extraction}

We select the review sentences that mention exactly one of the entities obtained in the previous 
\FT{stage} and pass these sentences through a BERT classifier to obtain votes based on the probabilities of whether the entity is a feature aspect, a product aspect, or not an aspect. We aggregate the votes for each of the entities and 
\FT{obtain a} list of extracted aspects.

\begin{figure}
	\centering
	\includegraphics[width=0.4\textwidth]{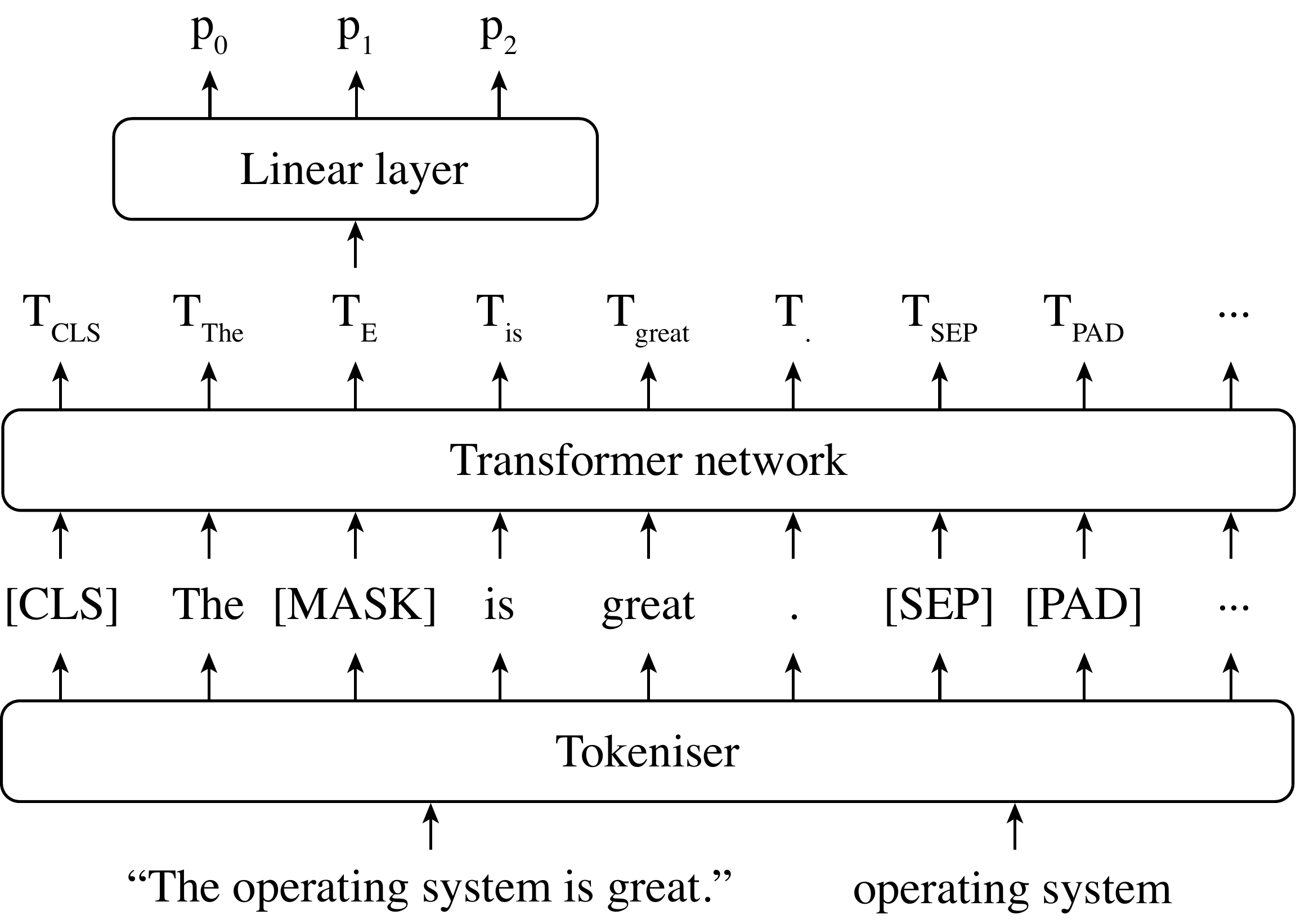}
	\caption{BERT for aspect extraction.}
	\label{fig:entityBERT}
\end{figure}

Figure \ref{fig:entityBERT} shows the architecture of the BERT classifier used for aspect extraction. The classifier takes as input a review sentence, as well as the entity we aim to classify as a feature aspect, product aspect, or non-aspect. The tokenisation step replaces the tokens associated with the entity (`operating' and `system') with a single mask token. The tokens are then passed through the transformer network and the input for classification is taken from the output position of the masked entity $T_E$. The linear layer is followed by a softmax operation, which outputs the probabilities $p_0$, $p_1$, and $p_2$ of the entity being a non-aspect, feature aspect, or product aspect, respectively.
Let $p_a = p_1 + p_2$ be the probability that an entity is any aspect of the product. For each entity, we take the mean of its $p_a$ votes \FT{(across different input review sentences where the entity occurs as the only entity)} and accept it as an aspect if the mean is above 0.65, a hyperparameter tuned through validation to strike a good balance between precision and recall. Similarly, an aspect is marked as a product aspect if the mean of its $p_2$ votes is above 0.45. Using the raw output probabilities from the network rather than binary votes allows us to bias the aggregate towards more certain predictions of the \FT{resulting} model.
We trained the model on the aspect extraction data 
\FT{(see Table \ref{tab:training_data})}, leaving 5\% of the data as a held-out \JO{testing} set. The final model was trained for 3 epochs with a batch size of 32 and Adam optimiser with standard cross entropy loss. The model was trained on a NVIDIA GeForce GTX 1080 GPU with 16GB RAM for 3 hours and 16 minutes. We obtain 82.37\% accuracy and 81.43\% macro $F_1$ on the testing set.

\subsection{Synonym Extraction}
\label{sec:synonym}
Reviewers can refer to the same aspect using  different terms (e.g. a \textit{hard drive} can also be referred to with the terms \textit{HDD} or \textit{storage}) that need to be grouped under the same aspect node 
\FT{when constructing the}  ontology tree.
However, terms such as \textit{hard drive} and \textit{storage} are not strict synonyms even if they are interchangeable within the review texts, rendering impossible the task of using a pre-existing synonym dictionary for grouping the aspects.
Since the terms are interchangeable within the review texts, we can again make use of the context of the terms to group terms with similar contexts into aspect \textit{synsets}. 
For this, we employ \textit{word embeddings} where the vectors of similar words are close to each other in the vector space. We obtain review-domain word embeddings by training a word2vec model on the review texts using the continuous bag-of-words (CBOW) architecture \FT{\cite{RefWorks:doc:5edbafdde4b0482c79eb8d95}} that predicts each word in the text corpus from a window of surrounding words.
\JNEW{We observed that a small window size of 4 \FTd{performed} best
. We hypothesise that this is because}
a small window size \JNEW{can} prevent
sibling aspects from being grouped together based on their association with their parent aspect, as shown in the following examples with the windows underlined:
\begin{center}
    \textit{I like this \textbf{screen} \underline{because of its sharp \textbf{resolution}}}

    \textit{I like this \textbf{screen} since \underline{it produces such vivid \textbf{colours}}}
\end{center}
Although both \textit{resolution} and \textit{colours} are mentioned in association with their parent aspect \textit{screen}, their nearby contexts are very different.

Once we have obtained the word embeddings for each of the aspects, we can use the \textit{relative cosine similarity} of the vectors to group them into synsets\FT{. Note that we use  
relative cosine similarity as this} is a more accurate measure for synonymy than cosine similarity \cite{RefWorks:doc:5eaebe76e4b098fe9e0217c2}.
The cosine similarity relative to the top $n$ most similar words between word embeddings $w_i$ and $w_j$ is calculated as
:

\(rcs_n(w_i,w_j) = \frac{cosine\_similarity(w_i,w_j)}{\sum_{w_c \in TOP_{i,n}}cosine\_similarity(w_i,w_c)},\)\\
where $TOP_{i,n}$ is the set of $n$ most similar words to $w_i$. If $rcs_{10}(w_i,w_j) > 0.10$, $w_j$ is more similar to $w_i$ than an arbitrary similar word from $TOP_{i,10}$\FT{: this} 
was shown to be a good indicator of synonymy \cite{RefWorks:doc:5eaebe76e4b098fe9e0217c2}.
We use the relative cosine similarity measure to construct a weighted graph of aspects to be used in the \textit{Equidistant Nodes Clustering} (ENC) method \cite{RefWorks:doc:5f46451de4b01aa839a177e7}, which was shown to obtain state-of-the-art performance in the synset induction task. We define a pair of aspects $(a_i, a_j)$ to be connected by an edge with weight $w_{i,j} = rcs_{10}(a_i,a_j) + rcs_{10}(a_j,a_i)$ if and only if $w_{i,j} \geq 0.21$. 
We use a maximum distance of $K=3$ and ranking function $rank\_wd$ \cite{RefWorks:doc:5f46451de4b01aa839a177e7} as parameters to ENC. 
We use the output clusters from ENC as our feature synsets, \FT{whereas the product synset is the set of product aspects obtained by the aspect extraction stage. } 


\subsection{Ontology Extraction}
\label{sec:ontology_extraction}

The aspect synsets obtained in the previous step will form the nodes 
of the ontology tree. We select the review sentences that mention a term from exactly two synsets and pass the sentences through a BERT classifier to obtain votes for 
relations between aspects. We then aggregate these votes within each synset to obtain a two-way relatedness measure between each synset pair that we use to create the ontology.

\begin{figure}
	\centering
	\includegraphics[width=0.4\textwidth]{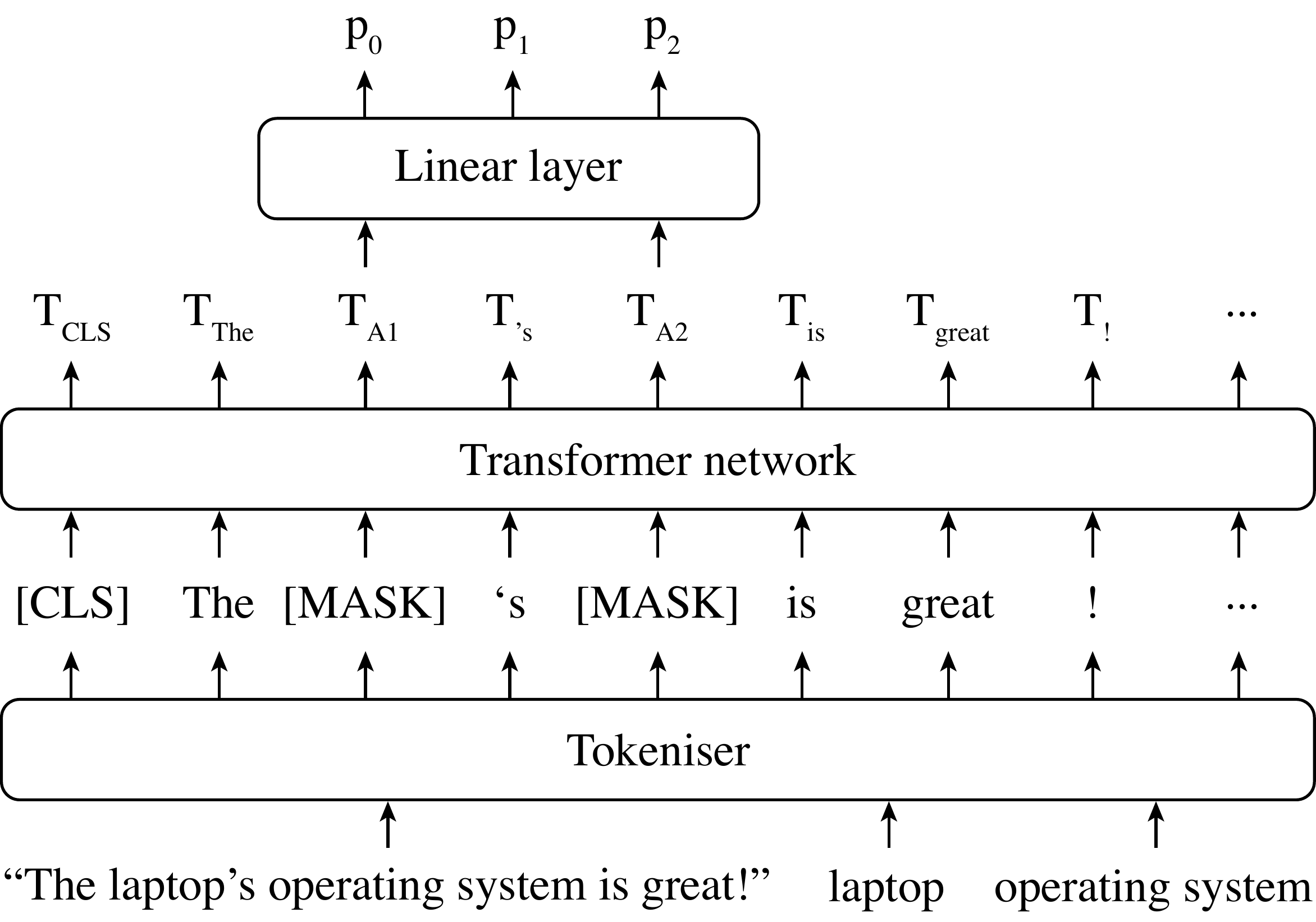}
	\caption{BERT for relation extraction.}
	\label{fig:relationBERT}
\end{figure}

\subsubsection{BERT for Relation Extraction.}

Figure \ref{fig:relationBERT} shows the architecture of the BERT classifier for relation extraction. The classifier takes as input a review sentence, as well as the two aspects $a_1$ and $a_2$ for which we wish to obtain one of three labels: 0 if $a_1$ and $a_2$ are not related, 1 if $a_2$ is a feature of $a_1$, and 2 if $a_1$ is a feature of $a_2$. The tokenisation step replaces the tokens associated with each of the aspects with a mask token. The tokens are passed through the transformer network and the input for the classification layer is taken from the output positions of the two masked aspects $T_{A1}$ and $T_{A2}$. The vectors are concatenated before being passed onto a linear classification layer with an output for each of the three labels. The classification layer is followed by 
softmax
which outputs the probabilities $p_0$, $p_1$, and $p_2$ of the three labels.
We trained the model on the relation extraction data 
\FT{(see Table \ref{tab:training_data})},
leaving 5\% of the data as a held-out \JO{testing} set. The final model was trained for 3 epochs with a batch size of 16 and Adam optimiser with standard cross entropy loss. The model was trained on a NVIDIA GeForce GTX 1080 GPU with 16GB RAM for 2 hours and 5 minutes. We obtain 79.73\% accuracy and 79.84\% macro $F_1$ on the testing set.

\subsubsection{Ontology Construction From Votes.}

Let $N$ be the number of synsets and $V \in \mathbb{N}^{N \times N}$ be a matrix where we accumulate the relation votes between each of the synsets. $V$ is initialised with zeroes, and for each vote $(p_0,p_1,p_2)$ on terms belonging to synsets $s_i$ and $s_j$ we 
add $p_1$ to 
the element $v_{j,i}$ of $V$ 
and $p_2$ to the element $v_{i,j}$ 
 of $V$. Upon completion, element $v_{i,j}$ of $V$ contains the sum of votes for $s_i$ being a feature of $s_j$ and element $v_{j,i}$ of $V$ contains the sum of votes for $s_j$ being a feature of $s_i$.

Let $n_{i,j}$ be the total number of input sentences to the relation classifier with terms from $s_i$ and $s_j$. Then %

\hspace*{3.2cm} $\bar{v}_{i,j} = \frac{v_{i,j}}{n_{i,j}}$\\
is the mean vote for $s_i$ being a feature of $s_j$. However, this measure has limitations as unrelated aspects may not co-appear in many sentences, leading to a high variance in calculating the mean. On the contrary, if $a_1$ is a feature of $a_2$, $a_1$ is likely to appear often in conjunction with $a_2$. We can use this observation to improve the accuracy of the relatedness measure.
Let $c_i$ be the total count of occurrences of terms from $s_i$ in the review texts. Then %

\hspace*{3.2cm} 
\(\tau_{i,j} = \frac{n_{i,j}}{c_i + c_j}\)\\
is a relative measure for how often aspects from $s_i$ and $s_j$ appear in conjunction. If we scale $\bar{v}_{i,j}$ by $\tau_{i,j}$, we obtain a more reliable measure for relatedness,

\hspace*{2cm} $r_{i,j} = \bar{v}_{i,j} \times \tau_{i,j} = \frac{v_{i,j}}{c_i + c_j}.$\\
We refer to matrix 
$(r_{i,j}) \in \mathbb{N}^{N \times N}$
as the \emph{relation matrix} $R$.

%



The product synset $s_p$ forms the root of the ontology tree. For each of the remaining synsets $s_i$, we calculate its super synset $\hat{s}_i$ using row $r_i$ of the relation matrix $R$ which contains the relatedness scores from $s_i$ to the other synsets. For example, the row corresponding to the synset of \textit{hands} for the product \textit{watch} could be as follows:

\begin{center}
 	\begin{tabular}{c c c c c c}
 	\textit{watch} & \textit{dial} & \textit{band} & \textit{battery} & \textit{hands} & \textit{quality} \\
 	\hline
 	0.120 & 0.144 & 0.021 & 0.041 & - & 0.037 \\ 
	\end{tabular}
\end{center}
Clearly, \textit{hands} appears to be a feature of \textit{dial} as the relatedness score for \textit{dial} is higher than for any other synset. We define $\hat{s_i}=s_j$ where $j = argmax(r_i)$, although other heuristics could be employed as well. Using the 
super synset relations, we expand the ontology starting from the root as shown in Algorithm \ref{alg:gettree}.

\begin{algorithm}
\DontPrintSemicolon
\SetKwInOut{Input}{input}
\SetAlgoLined

 \Input{R, synsets}
 nodes = \{$s$: Node($s$) for $s$ in synsets\}\;
 root = nodes[$s_p$]\;
 \For{\textnormal{$s$} in \textnormal{synsets} by descending order of \textnormal{R[$s$][$\hat{s}$]}}{
  \eIf{\textnormal{hasDescendant($s$, $\hat{s}$)}}{
   \tcp{loop in super relations}
   nodes[$s$].parent = root
   }{
   nodes[$s$].parent = nodes[$\hat{s}$]
  }
 }
 \Return root\;
 \caption{Ontology tree construction}
 \label{alg:gettree}
\end{algorithm}

\section{Evaluation} \label{sec:evaluation}

\FTd{To the best of our knowledge,  there is no publicly available gold standard dataset exclusively on product meronomies and no other method for extracting exclusively meronomies. Thus, for our evaluation, we opt for comparison with resources including meronomies and for methods extracting, amongst others, \emph{part of} relations. Concretely we evaluate our}
 method against a lexical database of semantic relations between words, WordNet \cite{RefWorks:doc:5f46c7b9e4b0bb83a698cd37}, and \JO{two} ontology learning framework\JO{s}, Text2Onto \cite{RefWorks:doc:5f46c8aae4b08553d3e6b982} \JO{and COMET \cite{Bosselut:19}}. 
We choose these baselines because
other \JNEW{ontology extraction} methods focus on extracting taxonomies as opposed to meronomies (i.e. \cite{KaramanolakisMD20,LuoLYBCWLYZ20,Shang:20, Shen:20, Kozareva:10}) or require extensive human intervention (i.e. \cite{RefWorks:doc:5f38122be4b0a8a5dc742bcd}), making them unsuitable for comparison. 
We considered also the approach of \cite{Konjengbam:18} for comparison with our approach, but the code 
\JNEW{is not} publicly available and, moreover,
the more promising results in \cite{Konjengbam:18} were obtained using manually labelled aspects.
\FTd{We conduct an evaluation with human annotators (Section~\ref{sec:human_eval}) as well as on a Q\&A dataset on Amazon products \cite{McAuleyY16,WanM16} (Section~\ref{sec:qa}). The latter evaluation aims at assessing how well terms in  ontologies (obtained with our method or by the baselines) match terms used in the Q\&A dataset, based on the intuition that a question about a product may be answered using information in the product's ontology.
}
Further, we evaluate the ability of our masked BERT models (i.e., 
aspect and relation extraction) to generalise \FT{(Section~\ref{sec:generalisation}).}

\begin{table}
    \centering
    \caption{Number of extracted terms $n_T$ and relations $n_R$.}
	\begin{tabular}{l| c c| c c| c c| c c}
	& \multicolumn{2}{c|}{\!Ours\!} & \multicolumn{2}{c|}{\!WordNet\!} & \multicolumn{2}{c|}{\!Text2Onto\!\!} & \multicolumn{2}{c}{\!COMET\!} \\ \hline
	\!\!Product\!\! & \!\!$n_T$\!\!\! & $n_R$\!\! & \!$n_T$\!\!\! & $n_R$\!\! & \!$n_T$\!\!\! & $n_R$\!\! & \!$n_T$\!\!\! & $n_R$\!\! \\
 	\hline
 	\!\!Watch\!\! & \!\!57\!\!\! & 36\!\! & \!\!82\!\!\! & 6\!\! & \!\!21\!\!\! & 20\!\! & \!\!25\!\!\! & 24\!\! \\
 	\!\!TV\!\! & \!\!50\!\!\! & 29\!\! & \!\!130\!\!\! & 19\!\! & \!\!22\!\!\! & 21\!\! & \!\!25\!\!\! & 24\!\! \\ 
 	\!\!Necklace\!\! & \!\!45\!\!\! & 20\!\! & \!\!13\!\!\! & 1\!\! & \!\!14\!\!\! & 13\!\! & \!\!25\!\!\! & 24\!\! \\ 
 	\!\!Mixer\!\! & \!\!42\!\!\! & 20\!\! & \!\!25\!\!\! & 6\!\! & \!\!25\!\!\! & 24\!\! & \!\!25\!\!\! & 24\!\! \\ 
 	\!\!Vid game\!\! & \!\!23\!\!\! & 16\!\! & \!\!3\!\!\! & 0\!\! & \!\!47\!\!\! & 46\!\! & \!\!25\!\!\! & 24\!\! \\ 
 	\hline
 	\!\!Total\!\! & \!\!217\!\!\! & 121\!\! & \!\!253\!\!\! & 32\!\! & \!\!129\!\!\! & 124\!\! & \!\!125\!\!\! & 120\!\! \\ 
	\end{tabular}
	\label{tab:ontology_counts}
\end{table}

\begin{figure}
	\centering
	\includegraphics[width=0.4\textwidth]{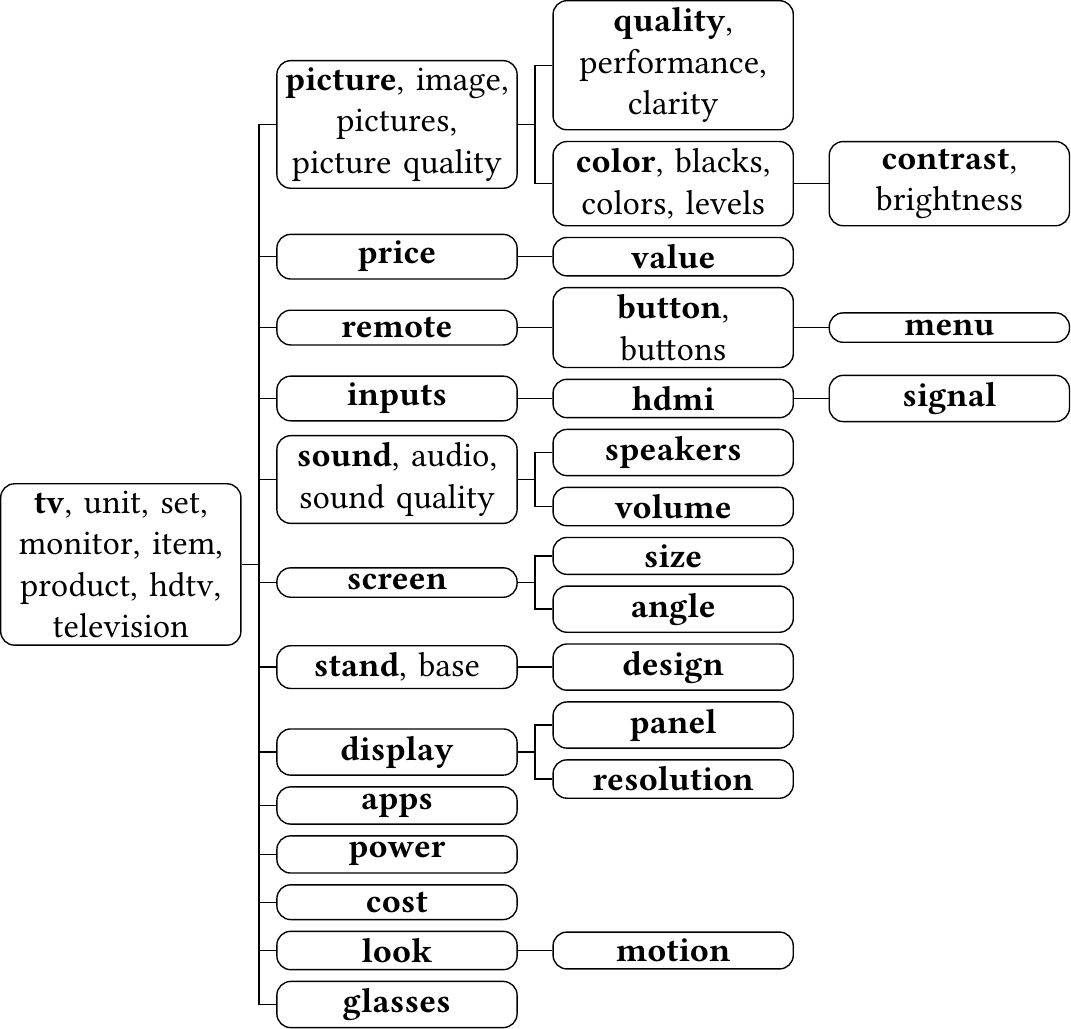}
	\caption{Ontology extracted by our method for \textit{television}. The highlighted terms were used in the human evaluation in Section \ref{sec:human_eval}.}
	\label{fig:tv_ontology}
\end{figure}

For the experiments \FTd{in Sections~\ref{sec:human_eval}--\ref{sec:qa}}, we evaluate the ontology \FTd{obtained} by our method \FTd{and the baselines} for  
five 
 randomly selected products which were not included in the training data\JNEW{, representative of 4 out of the 29 distinct Amazon categories in the dataset}: \textit{watches}, \textit{necklaces} \JNEW{(\FTd{both from category} Clothing Shoes and Jewelry)}, \textit{televisions} \JNEW{(Electronics)}, \textit{stand mixers} \JNEW{(Home and Kitchen)}, and \textit{video games} \JNEW{(Video Games)}. \JNEW{We have included both popular (\textit{televisions}) and niche (\textit{stand mixers}) products in attempt to cover the full spectrum of products.}%
 \footnote{For the experiments in Section~\ref{sec:generalisation}, to assess generalisability, we  \JNEW{include four more products for a total of} nine products.} %
For \FTd{these experiments, for} each product, we use 200,000 review texts as input to the ontology extractor, except for \textit{stand mixer}, for which we could only obtain 28,768 review texts due to it being a more niche category.
\FT{Our ontology is evaluated against} 
ontologies
 \FTd{(}for the five products\FTd{)} 
 from WordNet, Text2Onto, \JO{and COMET}, \FT{obtained} as follows. For WordNet, we construct the ontology top-down starting from the synsets for the product terms, \FTT{as follows. Let $s_p$ be a WordNet synset for one of the five products.  Then, our WordNet ontology includes a product synset $s'_p$ amounting to $s_p$ and all its direct hyponyms in WordNet. Further, a}  \JO{synset $s_f$} is considered a feature of \FT{ $s'_p$ } if \JO{$s_f$} is a meronym of any WordNet synset in  \JO{$s'_p$}.
  For Text2Onto, we use 2,000 review texts per product due to its imposed software limit discussed in \cite{Jiang:10}. We include in the product ontology each term related to the product term by the \textit{SubtopicOf} relation \FTT{in Text2Onto}. \JO{For COMET, we generate tuples from each of the product terms on ConceptNet's \textit{HasA} relation using the top-$k$ generation strategy. We choose $k=24$, which is closest to the average number of relations extracted by our method (24.2). The depth of the ontology for COMET is limited to one, as COMET can only query relations for one term at a time, and finding further sub-features without the context of the product term would lead to topic drift (e.g., the \textit{face} of a watch is very different from the \textit{face} of a human).}
Table \ref{tab:ontology_counts} shows the number of terms and relations in each extracted ontology. Figure \ref{fig:tv_ontology} shows the ontology extracted by our method for \textit{television}. 

 \begin{table*}
	\centering
	\caption{Precision \OC{(P)}, Relative \FT{(R.)} Recall \OC{(R)}, and $F_1$ scores for the four ontology extraction methods. For precision and recall, the total score is calculated across all relations. For $F_1$, the total score is macro $F_1$ \JO{across all five product categories}.}
	\begin{tabular}{l| c c c |c c c| c c c| c c c}
	& \multicolumn{3}{c|}{Ours} & \multicolumn{3}{c|}{WordNet} & \multicolumn{3}{c|}{Text2Onto} & \multicolumn{3}{c}{COMET} \\ \hline
	\!\!Product\!\! & P & \FT{R.} R & $F_1$ & P & \FT{R.} R & $F_1$ & P & \FT{R.} R & $F_1$ & P & \FT{R.} R & $F_1$ \\
 	\hline
 	\!\!Watch\!\! & 75.00 & \textbf{63.33} & \textbf{68.67} & \textbf{100.00} & 11.67 & 20.90 & 35.00 & 11.67 & 17.50 & 50.00 & 20.00 & 28.57 \\
 	\!\!TV\!\! & \textbf{82.76} & \textbf{50.67} & \textbf{61.80} & 68.42 & 24.00 & 36.25 & 38.09 & 13.33 & 20.15 & 54.17 & 17.33 & 26.80 \\ 
 	\!\!Necklace\!\! & 80.00 & \textbf{62.86} & \textbf{70.40} & \textbf{100.00} & 2.86 & 5.56 & 30.77 & 11.43 & 16.67 & 45.83 & 31.43 & 37.29 \\
 	\!\!Mixer\!\! & 80.00 & \textbf{50.00} & \textbf{61.54} & \textbf{100.00} & 25.00 & 40.00 & 33.33 & 20.45 & 25.35 & 16.67 & 9.09 & 11.76 \\
 	\!\!Vid game\!\! & \textbf{62.50} & \textbf{44.83} & \textbf{53.44} & - & 0.00 & 0.00 & 26.09 & 41.38 & 31.58 & 20.83 & 17.24 & 18.52 \\
 	\hline
 	\!\!Total\!\! & 76.86 & \textbf{54.92} & \textbf{63.24} & \textbf{81.25} & 15.16 & 20.48 & 31.45 & 17.21 & 22.21 & 37.50 & 18.44 & 24.52 \\
	\end{tabular}
	\label{tab:eval_scores}
\end{table*}

\subsection{Human Evaluation} \label{sec:human_eval}

\JNEW{\FTd{We are unaware of any} publicly available gold standard dataset for product meronomies, and annotating one is a difficult task prone to bias due to the boundless nature of parts that can be assigned to any given product. Thus, instead of evaluating against a gold standard meronomy,}
we present\FT{ed} each of the 397 meronym relations in the ontologies extracted by the four methods (see Table~\ref{tab:ontology_counts}) to three 
non-expert users who were given guidelines about the annotation process
and asked them to annotate the relation as either true or false in the general setting of product reviews. The setting is important as features such as \textit{price} may otherwise not be considered \FT{products' features}. 
\JO{For our method and WordNet, a synset can contain several terms, 
whereas the synsets for Text2Onto and COMET contain exactly one term. Thus, in this evaluation, we focus on the correctness of the relations, and only consider the most prominent term from each synset. 
In the case of our method, we present only the term with the highest prevalence in the review texts. In the case of WordNet, we present the first term in the hypernym synset.}
All three annotators agreed for 90.43\% of the relations. 
The inter-annotator agreement
, using Fleiss's $\kappa$ \cite{fleiss1971},  
was $\kappa = 0.872$.

\FT{We measured precision, \emph{relative} recall and 
$F_1$ for each product and overall for each of the methods (see Table~\ref{tab:eval_scores}), against human annotators.} %
We define the precision of an ontology as the aggregated precision of its individual relations, each given by  majority voting across the 
annotators. 
%
Our method achieves a total precision of 76.86\%.
%
WordNet obtains the highest total precision 
of 81.25\%, which may be expected as its knowledge has been manually annotated by human
\FT{s}. 
The sub-perfect precision for Wordnet's television ontology can be attributed to some highly scientific terms which the annotators did not recognise, such as \textit{electron gun}. 
However, WordNet extracted on average 6.4 relations for each product while our method extracted on average 24.2 relations \FT{(see Table~\ref{tab:ontology_counts})}.  This can be attributed to outdatedness\footnote{Last release in 2011: \url{https://wordnet.princeton.edu/news-0}}, although many of the products included are timeless (e.g., \textit{necklace}, \textit{watch}). 
%
The precision of our method is double that of Text2Onto
\FT{: this} is possibly due to the fact that the \textit{SubtopicOf} relation of Text2Onto is not limited to meronomies, \FT{which the annotators focused on.} \JO{COMET achieves slightly higher total precision than Text2Onto, but still performs significantly worse than our method, particularly on more niche categories such as \textit{stand mixers}.}

\FT{We define \emph{relative recall} as follows.} Let $T_{p,m}$ be the set of relations $(a_1, a_2)$ labelled as true by the majority of annotators in the ontology extracted by method $m$
for product $p$.
Assume the true ontology $O_p$ for a product $p$ is defined as:

\hspace*{1cm} \((a_1, a_2) \in O_{p} \iff \forall m. \ (a_1, a_2) \in T_{p,m}.\)\\
Using this assumption, we calculate the \textit{relative recall} $R_{p,m}$ for product $p$ and method $m$ as follows:

\hspace*{2.7cm} \(R_{p,m} = \frac{|T_{p,m}|}{|O_p|}.\)\\
Whilst $O_p$ may not represent the entire, possibly infinite, ontology for product $p$, $R_{p,m}$ is a good measure of relative recall for each of the methods as any addition to $O_p$ would decrease $R_{p,m}$ for all $m$. 
Table \ref{tab:eval_scores} shows the relative recall and the total relative recall 
\FT{for all methods}. Whilst WordNet obtained a
higher precision than our method, the relative recall of our method is three times 
higher than that of WordNet. 
Text2Onto and COMET perform slightly better in terms of recall, but still significantly worse than our method. This means that our method outperforms the other methods particularly in the breadth of its knowledge, as the others miss many key features of the products.
%
Finally, our method obtains the best macro-averaged $F_1$ score, with the $F_1$ score for \textit{stand mixer} 
comparable to the scores 
for the other 
\FT{products} despite using less data, showing that our method is effective even for products with 
relatively little data.

\subsection{Q\&A Dataset Evaluation}
\label{sec:qa}

 \FTd{We} further evaluate our method using an \JNEW{automatic method exploiting an} Amazon Q\&A dataset\footnote{http://jmcauley.ucsd.edu/data/amazon/qa/} which contains questions and answers about products. The Q\&A instances often concern the
aspects of the given product as in the following example: 
\begin{center}
    Q: \textit{What are the dimensions of this \textbf{TV}?} \\
    A: \textit{The \textbf{screen} is 38.5 inches.}
\end{center}

\begin{table}
	\centering
	\caption{Results for the Q\&A dataset where $p_{a}$ is the percentage of Q\&A instances that mention at least one aspect from the ontology and $p_{r}$ is the percentage of instances containing a parent/child relation from the ontology in the question/answer, respectively.}
	\begin{tabular}{l| c c| c c| c c| c c}
	& \multicolumn{2}{c|}{\!\!Ours\!\!} & \multicolumn{2}{c|}{\!\!WordNet\!\!} & \multicolumn{2}{c|}{\!\!Text2Onto\!\!} & \multicolumn{2}{c}{\!\!COMET\!\!} \\ \hline
	\!\!Product\!\! &  \!\!$p_{a}$ &  $p_{r}$ &  \!\!$p_{a}$ &  \!$p_{r}$ &  \!\!$p_{a}$ &  $p_{r}$ &  \!\!$p_{a}$ &  $p_{r}$\!\! \\
 	\hline
 	\!\!Watch\!\! & \!\!\textbf{81.7}\!\!\! & \textbf{23.2}\!\! & \!\!63.4\!\!\!\! & 0.0\!\! & \!\!64.6\!\!\!\! & 6.1\!\! & \!\!70.7\!\!\!\! & 11\!\! \\
 	\!\!TV\!\! & \!\!\textbf{86.5}\!\!\! & \textbf{23.7}\!\! & \!\!73.9\!\!\!\! & 23.6\!\! & \!\!64.2\!\!\!\! & 4.7\!\! & \!\!31.6\!\!\!\! & 0.2\!\! \\ 
 	\!\!Necklace\!\! & \!\!\textbf{60.0}\!\!\! & \textbf{6.1}\!\! & \!\!27.7\!\!\!\! & 0.0\!\! & \!\!30.8\!\!\!\!\! & 1.5\!\! & \!\!30.8\!\!\! & 0.0\!\! \\ 
 	\!\!Mixer\!\! & \!\!\textbf{78.5}\!\!\! & \textbf{23.2}\!\! & \!\!42.2\!\!\!\! & 0.4\!\! & \!\!54.0\!\!\!\! & 3.4\!\! & \!\!12.2\!\!\!\! & 0.0\!\! \\
 	\!\!Vid Game\!\! & \!\!\textbf{43.6}\!\!\! & 2.7\!\! & \!\!0.3\!\!\!\! & 0.0\!\! & \!\!43.5\!\!\!\! & \textbf{3.4}\!\! & \!\!6.9\!\!\!\! & 0.0\!\! \\
 	\hline
 	\!\!Average\!\! & \!\!\textbf{70.1}\!\!\! & \textbf{15.8}\!\! & \!\!41.5\!\!\!\! & 4.8\!\! & \!\!51.4\!\!\!\! & 3.8\!\! & \!\!30.4\!\!\!\! & 2.2\!\! \\
	\end{tabular}
	\label{tab:qa_eval}
\end{table}

We evaluate a product ontology by determining how well it corresponds to the Q\&A data for that product category 
using an automatic method that matches terms in the ontology with the Q\&A texts.
Specifically, we measure the percentage $p_a$ of Q\&A instances that mention at least one aspect from the ontology. To evaluate the relations, we measure the percentage of instances containing a parent/child relation from the ontology in the question/answer, respectively. 
Table \ref{tab:qa_eval} shows the results for the previously used categories for testing: \textit{watches} (with 82 \FT{Q\&A} instances), \textit{televisions} (4185), \textit{necklaces} (65), \textit{stand mixers} (237), and \textit{video games} (9660).
Our method outperforms the other 
methods in all but one category, \textit{video games}, where Text2Onto obtains a similar $p_a$ 
but a higher $p_r$
. This is in line with the human evaluation 
where \textit{video games} was the \JNEW{worst-}performing category, in terms of $F_1$
, for our method and the best for Text2Onto. On average, our method performs better than the 
other
s, 
in particular for relations, where it performs three times better. \JO{While COMET outperformed WordNet and Text2Onto in the human evaluation (macro $F_1$), it obtains the lowest scores in this task suggesting that it is less applicable in
a real\JNEW{-}world setting.} \JNEW{This experiment shows that the ontologies extracted by our method contain common-sense knowledge that can be applied in a real-world Q\&A setting. It also shows that product aspects are at the center of e-commerce Q\&A, underlining the overlooked potential of meronomies in structuring and explaining Q\&A data.}

\subsection{Generalisation Evaluation} \label{sec:generalisation}

Finally, we evaluate the ability of our masked BERT methods, the aspect and relation classifiers, to generalise  
for unseen products \JNEW{despite being trained on masked data for only a few product categories}.
%
\JNEW{In order to do this, we train the classifiers on incremental datasets of increasing product variety and evaluate them on unseen products. We wish to ascertain the degree of product variety in the training set needed to achieve good performance on unseen products; generalisation can be deemed to happen only if the training step does not require a huge variety of products for the model to perform well on unseen products.}
\JNEW{We draw the product categories used for training and evaluation from} the \JNEW{set of five product categories used in the previous experiments and an additional four product categories for which data has been similarly annotated, for a total of nine product categories}: $S=$ \{\textit{digital cameras}, \textit{backpacks}, \textit{laptops}, \textit{acoustic guitars}, \textit{cardigans}, \textit{watches}, \textit{necklaces}, \textit{video games}, \textit{headphones}\}, where each category has 42,313 instances for the aspect classifier and 21,131 instances for the relation classifier. 
All models \FT{discussed below} were trained using the configurations given in Sections \ref{sec:feature_extraction} and \ref{sec:ontology_extraction}.

\begin{figure}
    \centering
    \includegraphics[width=0.39\textwidth]{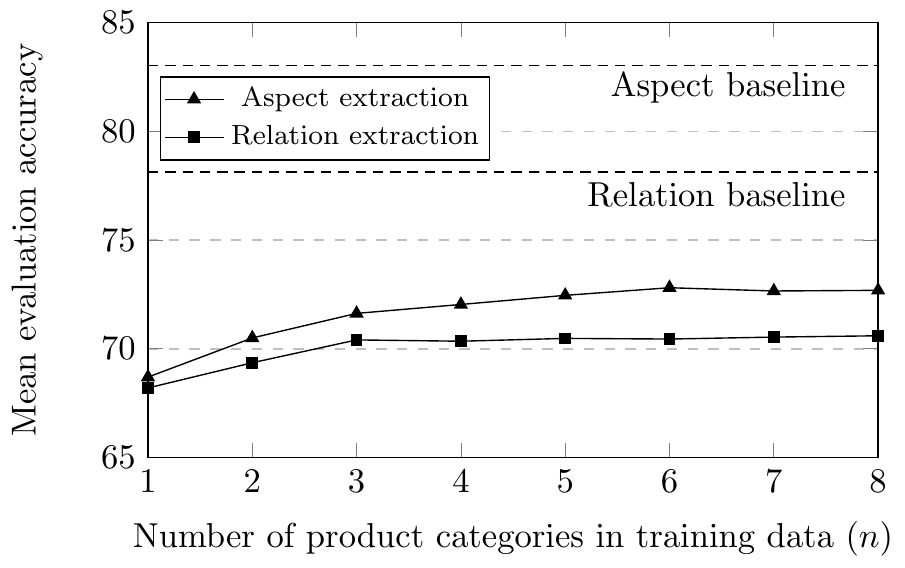}
    \caption{Mean accuracies on the $U_{n+1}$ test set for masked BERT models trained on $T_n
    $
    . The mean in-domain accuracies of the models are shown as dashed lines.}
    \label{fig:n_accuracies}
\end{figure}

\begin{table}
	\centering
	\caption{Error values when using the method in \cite{RefWorks:doc:5e1717fde4b078c95934fc03} with the ontology extraction methods.}
	\begin{tabular}{l| c c| c c| c c| c c}
	& \multicolumn{2}{c|}{Ours} & \multicolumn{2}{c|}{WordNet} & \multicolumn{2}{c|}{Text2Onto} & \multicolumn{2}{c}{\!COMET} \\ \hline
	\!\!Product\!\! & \!\!MAE\!\! & \!\!RMSE\!\! & \!\!MAE\!\! & \!\!RMSE\!\! & \!\!MAE\!\! & \!\!RMSE\!\! & \!\!MAE\!\! & \!\!RMSE\!\!\! \\
 	\hline
 	\!\!Watch\!\! & \!\!\textbf{6.7}\!\! & \!\!\!\textbf{8.5}\!\!\! & \!\!14.8\!\!\! & \!\!\!15.3\!\!\! & \!\!13.9\!\!\! & \!\!\!14.5\!\!\! & \!\!12.0\!\!\! & \!\!\!12.6\!\!\! \\
 	\!\!TV\!\! & \!\!\textbf{3.4}\!\! & \!\!\!\textbf{6.7}\!\!\! & \!\!6.6\!\!\! & \!\!\!8.5\!\!\! & \!\!23.2\!\!\! & \!\!\!23.6\!\!\! & \!\!28.6\!\!\! & \!\!\!28.8\!\!\! \\ 
 	\!\!Necklace\!\! & \!\!\textbf{8.8}\!\! & \!\!\!\textbf{10.9}\!\!\! & \!\!28.3\!\!\! & \!\!\!29.0\!\!\! & \!\!29.1\!\!\! & \!\!\!29.4\!\!\! & \!\!27.2\!\!\! & \!\!\!28.6\!\!\! \\ 
 	\!\!Mixer\!\! & \!\!22.7\!\! & \!\!\!23.4\!\!\! & \!\!28.9\!\!\! & \!\!\!29.4\!\!\! & \!\!\textbf{21.6}\!\!\! & \textbf{22.2} & \!\!35.2\!\!\! & \!\!\!35.6\!\!\! \\
 	\!\!Vid game\!\! & \!\!12.6\!\! & \!\!\!14.6\!\!\! & \!\!33.2\!\!\! & \!\!\!34.9\!\!\! & \!\!\textbf{10.4}\!\!\! & \!\!\!\textbf{11.8}\!\!\! & \!\!21.8\!\!\! & \!\!\!24.2\!\!\! \\
 	\hline
 	 \!\!Average\!\! & \!\!\textbf{10.8}\!\! & \!\!\!\textbf{12.8}\!\!\! & \!\!22.4\!\!\! & \!\!\!23.4\!\!\! & \!\!19.6\!\!\! & \!\!\!20.3\!\!\! & \!\!25.0\!\!\! & \!\!\!26.0\!\!\! \\
	\end{tabular}
	\label{tab:strength_eval}
\end{table}

The generalisation evaluation is defined as follows: we train the aspect and relation classifiers with eight \JO{equally large} incremental datasets $T_1 ... T_8$, where $T_n$ includes reviews from $n$ different categories of products selected from $S$. 
For each $T_n$, we evaluate the trained model on an \emph{unseen} dataset 
$U_{n+1}$
containing reviews from a single product category randomly selected from $S$ \JO{but not included in $T_n$}. We run this experiment five times, randomly selecting 
$(T_n, U_{n+1})$, 
and average the results. \FT{As} baselines, we provide \OC{the average of} in-domain baseline scores obtained by training and evaluating the aspect and relation classifiers on five randomly selected $(T_1, U_1)$ where $T_1$ and $U_1$ contain reviews from the same product category. 

Figure \ref{fig:n_accuracies} shows the mean evaluation accuracies for each of the two classifiers trained on $T_1 ... T_8$ as well as the mean in-domain baseline accuracies. 
The accuracies for both entity and relation extraction increase significantly when trained with reviews for three products instead of a single product, after which the accuracy for relation extraction stays constant around 8\% below the in-domain accuracy. The aspect extraction accuracy improves slightly until five products used in training, after which it stays somewhat constant at 10\% below the in-domain accuracy. 
It is somewhat expected 
for the accuracy to increase as the number of products used for training increases, since using a classifier on a product-specific dataset forces the classifier to learn product-specific features which may not apply to an unseen product domain.   However, it is interesting to note that training the classifier with just 3-5 products is enough to raise its accuracy to its domain-independent optimum. The domain-specific classifier's advantage of around 8-10\% over the domain-independent classifier  can be attributed to various domain-specific unmasked context features that the classifier learns to take advantage of, such as 
adjectives like \textit{swiss} or \textit{waterproof} for \textit{watch}.

\section{Case Study: a Product Recommender System} \label{sec:applications}

Ontologies have applications in many settings, including  recommender systems (e.g., as in   \cite{Iswari:19,Xu:20}) and search engines (e.g., as in \cite{Ramkumar:14,Fazzinga:10}). 
In this section, we focus on the former class of applications and show how our method to extract ontologies (in the form of meronomies) can fruitfully support a product recommender system, competitively with  the methods evaluated in  Section \ref{sec:evaluation}. Specifically, 
we define a product recommender system adapting the method of \cite{RefWorks:doc:5e1717fde4b078c95934fc03}.
Whereas the latter 
uses a (part hand-crafted, part automatically mined) movie ontology
and reviews/aggregated scores for movies from 
Rotten Tomatoes\footnote{\url{https://www.rottentomatoes.com/}}, we use the product ontology obtained by our method (and used for the evaluation in Section~\ref{sec:evaluation}) and Amazon's textual reviews/aggregated scores
for the products described in the ontology.
Here, as in \cite{RefWorks:doc:5e1717fde4b078c95934fc03}, the aggregated scores available from the web sites providing them are used as ``ground truth'' for the evaluation of the recommendations, which are obtained fully from the reviews, using the ontologies, as follows. 

First (positive or negative) votes on aspects of relevant products, from the given ontology, are drawn automatically from the reviews, by automatically identifying the (positive or negative, respectively) sentiment of excerpts of the reviews concerning those aspects \JNEW{using the state-of-the-art method of TD-BERT \cite{Gao:19} trained on SemEval-2014 Task 4 data for aspect-based sentiment analysis \cite{Pontiki:14}}. Then, the votes are used to obtain an \emph{argument graph}, whereby (excerpts about) aspects of the ontology form arguments supporting or attacking other aspects or the (``goodness'' of the) product.    
Finally, 
the argument graph is evaluated dialectically, in the spirit of computational argumentation, to obtain a numerical score for the product (see \cite{RefWorks:doc:5e1717fde4b078c95934fc03} for details).
This score 
is then compared, for evaluation, to the ``ground truth'' aggregated score 
(from Rotten Tomatoes in \cite{RefWorks:doc:5e1717fde4b078c95934fc03} and from Amazon here, amounting to the average star ratings of the products therein, scaled linearly from 0-5 to 0-100).
The intuition is that the computed score gives a measure of the dialectical support products  receive from the opinions expressed in the reviews, along the dimensions identified by the ontology, and thus this score should be comparable to the aggregated star ratings reviewers give the products.
%


Table \ref{tab:strength_eval} shows the \OC{mean average error} (MAE) and \OC{root mean square error} (RMSE) for each ontology extraction method and product category. Here, we compute scores for the 20 most reviewed products in each testing category
. 
%
%
On average, our method performs the best as it obtains the lowest average error scores. However, for \textit{stand mixer} and \textit{video game}, Text2Onto obtains lower error scores compared to our method. This may be 
because
Text2Onto obtained its highest $F_1$ scores in these categories.
This experiment suggests that our 
method may be well suited to support applications built from review texts.
Note that, by leading to ``dialectical'' scores better matching the aggregate scores on Amazon, our method could support more meaningful explanations of the scores, 
allowing users to decide whether a product may be useful to them depending on its features. 
For illustration, 
a user may be more interested in the \textit{fabric} of a product than its \textit{price}, information that cannot be easily obtained without reading all reviews of that product but that our method, in combination with the approach of \cite{RefWorks:doc:5e1717fde4b078c95934fc03}, could provide. 
We leave the exploration of this line of research as future work.

\section{Conclusions} \label{sec:conclusion}

We proposed a novel method using masked BERT to identify aspects and relations between them to construct ontologies, in the form of meronomies,  from textual product reviews. Our method uses distantly supervised learning from a small, manually constructed ontology for a handful of products to automatically annotate large amounts of text. We showed that our method generalises well and outperforms, in different evaluation settings, a hand-crafted ontology and existing methods for extracting ontologies. Our experiments with a Q\&A dataset and a downstream recommender system task indicate that meronomies have great uncharted potential as specialised forms of ontologies.

As future work, we envisage a wide variety of potential applications of our method, for example in conjunction with attribute value extraction methods, e.g. \cite{Wang:20}, a similarly generic method, to obtain highly structured information about any product from plain text. Furthermore, our method is not limited to product reviews only. Future work includes testing the method in other applications, such as the biological domain, as well as identifying other types of relations between aspects to support a broader class of ontologies than the meronomies we focused on.

\bibliographystyle{ACM-Reference-Format}
\bibliography{ontology}


\end{document}